\documentclass[alpha-refs]{wiley-article}

\usepackage{color}
\usepackage{ulem}

\usepackage{siunitx}

\usepackage{float}
\usepackage{pifont}
\usepackage{bm}
\usepackage{textcomp}
\usepackage{grffile}
\usepackage[final]{pdfpages}
\usepackage{pdflscape}
\usepackage{pifont}
\usepackage{adjustbox}
\usepackage{bbm}

\usepackage{colortbl}
\usepackage{multirow}
\usepackage{hhline}
\usepackage{nicematrix}
\usepackage{array}
\usepackage{xparse}
\usepackage{xifthen}
\usepackage{hyperref}
\usepackage{cleveref}
\usepackage{ifdraft}
\usepackage[inline]{enumitem}

\usepackage{mwe}

\usepackage[parfill]{parskip}
\usepackage{marvosym}
\usepackage{tasks}

\DeclareMathOperator*{\argmax}{arg\,max}
\DeclareMathOperator*{\argmin}{arg\,min}

\newcommand{\MT}{\lfloor M\cdot\theta  \rceil}

\newcommand{\EE}{\mathbb{E}}
\newcommand{\PP}{\mathbb{P}}
\newcommand{\RR}{\mathbb{R}}
\newcommand{\NN}{\mathbb{N}}

\newcommand{\TAC}[1]{\sigma_{#1}}

\newcommand{\UD}[1]{\sim \mathcal{U}\left (#1\right )}

\newcommand{\permvec}[2]{#1_{#2}}
\newcommand{\permmat}[2]{#1_{#2}}

\newcommand{\dequal}{\stackrel{d}{=}}

\newcommand{\hmii}{\mathcal{H}^{i.i.}_M}

\newcommand{\uar}{u.a.r.}

\papertype{Original Article}

\title{The Optimal Input-Independent Baseline for Binary Classification: The Dutch Draw}

\author[1]{Joris Pries}
\author[1]{Etienne van de Bijl}
\author[1]{Jan Klein}
\author[2]{Sandjai Bhulai}
\author[1,2]{Rob van der Mei}

\affil[1]{Department of Stochastics, Centrum Wiskunde \& Informatica, Amsterdam, North Holland, 1098 XG, Netherlands}

\affil[2]{Department of Mathematics, Vrije Universiteit, Amsterdam, North Holland, 1081 HV, Netherlands}

\corraddress{Joris Pries, Department of Stochastics, Centrum Wiskunde \& Informatica, Amsterdam, North Holland, 1098 XG, Netherlands}
\corremail{joris.pries@cwi.nl}

\fundinginfo{No additional funding}

\runningauthor{Joris Pries et al.}

\begin{document}

\maketitle

\begin{abstract}
Before any binary classification model is taken into practice, it is important to validate its performance on a proper test set. Without a frame of reference given by a baseline method, it is impossible to determine if a score is `good' or `bad'. The goal of this paper is to examine all baseline methods that are independent of feature values and determine which model is the `best' and why. By identifying which baseline models are optimal, a crucial selection decision in the evaluation process is simplified. We prove that the recently proposed \emph{Dutch Draw baseline} is the best \emph{input-independent} classifier (independent of feature values) for all \emph{positional-invariant} measures (independent of sequence order) assuming that the samples are randomly shuffled. This means that the \emph{Dutch Draw baseline} is the optimal baseline under these intuitive requirements and should therefore be used in practice.

\end{abstract}

\section{Introduction} \label{sec: introduction}
A \emph{binary classification} model is trying to answer the following question: Should the instance be labeled as zero or one? This question might seem simple, but there are many practical applications for binary classification, ranging from predicting confirmed COVID-19 cases \citep{Pirouz}, detecting malicious intrusions \citep{Longjie} to determining if a runner is fatigued or not \citep{Buckley}. Whenever a classification model is developed for a practical application, it is important to validate the performance on a test set. However, a baseline is necessary to put the achieved performance in perspective. Without this frame of reference, only partial conclusions can be drawn from the results. An accuracy of 0.9 indicates that 90\% of all predictions are correct. But it could be that the model actually did not learn anything and such a high accuracy can already be achieved by predicting only zeros. To put the performance in perspective, it should therefore be compared with some meaningful benchmark method, preferably with a state-of-the-art model.

Nevertheless, many state-of-the-art methods are very problem-specific. They can rapidly change and often involve many fine-tuned parameters. Thus, as a necessary additional check in the development process, \Citet{Dutchdraw} plead for a supplementary baseline that is \emph{general}, \emph{simple}, and \emph{informative}. This baseline should test if the new model truly performs better than a simple model. It should be considered a major warning sign when a model is outperformed by e.g., a weighted coin flip. The binary classification model can use information about the feature values of a sample, yet it is outperformed by a model that does not even consider these values. Is the model then actually learning something productive?

A theoretical approach for binary classification is proposed in \Citep{Dutchdraw} based on \emph{Dutch Draw classifiers}. Such a classifier draws uniformly at random (\uar{}) a subset out of all samples, and labels these 1, and the rest 0. The size of the drawn subset is optimized to obtain the optimal expected performance, which is the \emph{Dutch Draw baseline}. For most commonly used performance measures, a closed-form expression is given \Citep{Dutchdraw}.

However, there are infinitely many ways to devise a baseline method. We only investigate prediction models that do not take any information from the features into account, as this will result in a more general and simple baseline. We call these models \emph{input-independent}. Irrespective of the input, the way that such a model predicts remains the same. Any newly developed model should at least beat the performance of these kinds of models, as an \emph{input-independent} model cannot exploit patterns in the data to predict the labels more accurately. However, sometimes a model can get lucky by accidentally predicting the labels perfectly for a specific order of the labels. The order of the samples should not influence the `optimality' of a model. This is why we introduce the notion of \emph{permutation-optimality}. Furthermore, the order of the samples should not change the outcome of the performance measure (\emph{positional-invariant}). This is not a strict condition, as most commonly used measures have this property. Under these restrictions, we prove that the \emph{Dutch Draw baseline} is \emph{permutation-optimal} out of all \emph{input-independent} classifiers for any \emph{positional-invariant} measure.

To summarize, in this paper we:
\begin{itemize}
    \item determine natural requirements for a general, informative and simple baseline;
    \item prove that the Dutch Draw baseline is the optimal baseline under these requirements.
\end{itemize}

These contributions improve the evaluation process of any new binary classification method.

The remainder of this paper is organized as follows. First, the necessary preliminaries and notations are discussed in Section~\ref{sec: preliminaries}. Next, in Section~\ref{sec: essential conditions} we determine requirements for a general, simple and informative baseline. Furthermore, we formally define what optimality entails under these requirements. In Section~\ref{sec: dutch draw classifier}, an alternative definition for the \emph{Dutch Draw classifiers} is given, which is necessary for the main proof. In Section~\ref{sec: proof}, we prove that the Dutch Draw baseline is optimal. Finally, Section~\ref{sec: discussion and conclusion} summarizes the general findings and discusses possible future research opportunities.

\section{Preliminaries} \label{sec: preliminaries}
Next, we introduce some concepts and notations to lay the foundation for the main proof. First, \emph{binary classifiers} (Section~\ref{sec: binary classifiers}) and \emph{performance measures} (Section~\ref{sec: binary classification measures}) for binary classification are discussed. Then, properties of \emph{permutations} are examined in Section~\ref{sec: permutations}, which will play a crucial role in the proof of the main result.

\subsection{Binary classifiers} \label{sec: binary classifiers}
To find a good baseline for a \emph{binary classification model}, we first have to discuss what a \emph{binary classifier} actually is. To this end, let $\mathcal{X}$ be the feature space (think e.g., $\RR^d$). Normally, a binary classifier is defined as a function ${h: \mathcal{X} \to \{0,1\}}$ that maps feature values to zero or one. However, this classifier only classifies one sample at a time. Instead, we are interested in classifiers that classify \emph{multiple} samples \emph{simultaneously}:
\begin{align*}
    h_M &: \mathcal{X} ^ M \to \{0,1\} ^ M,
\end{align*}
where $M \in \NN_{>0}$ denotes the number of samples that are classified. This gives classifiers the ability to precisely predict $k$ out of $M$ samples positive. Note that a \emph{single sample} classifier $h$ can simply be extended to classify $M$ samples \emph{simultaneously} by applying the classifier for each sample individually: \begin{align*}
h_M: (x_1, \dots, x_M) \mapsto \left (h(x_1), \dots, h(x_M) \right ).
\end{align*}
Let $\mathcal{H}_M = \{ h_M: \mathcal{X} ^ M \to \{0,1\} ^ M\}$ be the set of all binary classifiers that classify $M$ samples at the same time.

\subsubsection*{Example of a binary classifier} \label{sec: example binary classifier}
\newcommand{\coin}{h_{\text{coin}}}
\newcommand{\coinn}{h_{\text{coin}}^{\text{single}}}
An example of a binary classifier is a \emph{coin toss}, where each sample is classified by throwing a coin and determining on which side it lands. Let $\theta \in [0,1]$ be the probability that the coin lands head, and $1-\theta$ for tails. Assuming that head and tails are classified by 1 and 0 respectively, we get: \begin{align*}
    \coinn(\cdot) := \left \{ \begin{array}{ll}
        1 & \text{ with probability } \theta,\\
        0 & \text{ with probability } 1-\theta.\\
    \end{array}
    \right .
\end{align*}

Classifying $M$ samples by repeatedly throwing coins can be achieved by: \begin{align*}
    \coin : (x_1, \dots, x_M) \mapsto \left ( \coinn(x_1), \dots, \coinn(x_M)  \right ).
\end{align*}

\subsection{Performance measures for binary classification}\label{sec: binary classification measures}
To assess the effectiveness of a binary classification model, it is necessary to choose a \emph{performance measure}, which quantifies how much the \emph{predicted} labels agree with the \emph{actual} labels. Namely, each sample indexed by $i$ has feature values ${\mathbf{x}_i \in \mathcal{X}}$ and a corresponding label ${y_i\in \{0, 1\}}.$ Let ${\mathbf{X} := \left (\mathbf{x}_1 \dots \mathbf{x}_M \right ) \in \mathcal{X}^M}$ be the combined feature values of $M$ samples. Furthermore, let ${\mathbf{Y}= (y_1, \dots, y_M)}$ denote the corresponding labels. A performance measure for binary classification is then defined as ${\mu : \{0,1\} ^M \times \{0,1\} ^M \to \RR}$, where the first entry of $\mu$ is the predictions made by the classifier and the second entry is the corresponding labels. The performance of classifier $h_M$ can now be written as: $\mu(h_M(\mathbf{X}) , \mathbf{Y}).$

\subsubsection*{Example of a performance measure}\label{sec: example binary classification measure}
An example of a performance measure for binary classification is \emph{accuracy} ($\mu_{\text{acc}}$). It is defined as the total number of correctly classified samples divided by the total number of samples. For any ${h_M(\mathbf{X})= (\hat{y}_1,\dots, \hat{y}_M)\in \{0,1\} ^M}$ and $ {\mathbf{Y}=(y_1,\dots,y_M) \in \{0,1\} ^M}$, it holds that \begin{align*}
    \mu_{\text{acc}}\left (h_M(\mathbf{X}) , \mathbf{Y}\right ) &= \frac{\sum_{i = 1}^{M} \mathbbm{1}_{\{\hat{y}_i = y_i\}} }{M}. %
\end{align*}

\subsubsection*{Undefined cases} \label{sec: undefined measure}
Some measures are undefined for specific combinations of $h_M(\mathbf{X})$ and $\mathbf{Y}$. Take for example the \emph{true positive rate} \citep{Tharwat}, which is the number of correctly \emph{predicted} positives divided by the total number of \emph{actual} positives. When there are no actual positives, the measure is ill-defined, as it divides by zero. Less obvious, the measure \emph{negative predictive value} \citep{Tharwat} is undefined when no negatives are predicted, as it is defined as the number of correctly predicted negatives divided by the total number of predicted negatives. Defining $\frac{x}{0}:=0$ for all $x\in \mathbb{R}$ will solve many undefined issues. However, this can make it desirable for a classifier to always predict labels that lead to a previously undefined measure in order to minimize the measure. Therefore, we redefine $\mu$ from now on for every ${\hat{\mathbf{Y}}, \mathbf{Y} \in \{0,1\}^M}$ to be equal to a constant $C_{\text{undef}}$, when ${\mu(\hat{\mathbf{Y}}, \mathbf{Y})}$ was undefined. We make a distinction for each objective (maximizing/minimizing). Let \begin{align*} 
    C_{\text{undef}} := \left \{ \begin{array}{ll}
        \max_{\hat{\mathbf{Y}}, \mathbf{Y} \in \{0,1\}^M} \left \{\mu(\hat{\mathbf{Y}}, \mathbf{Y}) \right \} & \text{if minimizing},\\
        \min_{\hat{\mathbf{Y}}, \mathbf{Y} \in \{0,1\}^M} \left \{\mu(\hat{\mathbf{Y}}, \mathbf{Y}) \right \}  & \text{if maximizing}.
    \end{array} \right .
\end{align*}
It is therefore always disadvantageous for a classifier to predict a previously undefined case. By defining $C_{\text{undef}}$ in this way, we do not have to omit such classifiers from our analysis.

\subsection{Permutations} \label{sec: permutations}
To determine which binary classifier is considered to be the `best', we define \emph{permutation-optimality} in Section~\ref{sec: permutation-optimality}, which uses \emph{permutations} to define `optimality'. In this section, we examine properties of permutations that are used in the main proof (see Section~\ref{sec: proof}). A permutation is a \emph{bijective} function from a set to itself \citep{Dixon}. This means that a permutation is not a reordered list; it is a function that determines where each element should be rearranged to.

Let $S_{M}$ denote the set of all permutations of a set of size $M$, also called the \emph{symmetric group}. More formally,
\begin{align*}
    S_{M}:= \left \{\pi : \{1,\dots, M \} \to \{1,\dots, M \} \text{ s.t. } \{\pi(i)\}_{i = 1}^{M} = \{1,\dots, M\} \right \}.
\end{align*}

\subsubsection*{Example of symmetric group}
Using the Cauchy one-line notation \citep{Cauchy}, all possible permutations of three elements are given by
\begin{align*}
    \begin{array}{lll}
      \begin{pmatrix}
          1 & 2 & 3
      \end{pmatrix}, &
      \begin{pmatrix}
        1 & 3 & 2
    \end{pmatrix},&
    \begin{pmatrix}
        2 & 1 & 3
    \end{pmatrix},\\
    & & \\
    \begin{pmatrix}
        2 & 3 & 1
    \end{pmatrix},&
    \begin{pmatrix}
      3 & 1 & 2
  \end{pmatrix},&
  \begin{pmatrix}
      3 & 2 & 1
  \end{pmatrix}.\\
    \end{array}
\end{align*}
The permutation $\begin{pmatrix}
    2 & 3 & 1
\end{pmatrix}$ sends the first element to the second position, the second element to the third position and the third element to the first position.

\subsubsection*{Sample-wise permutations}
To apply permutations to a matrix, we discuss \emph{sample-wise permutations}. For every $M\times K$ dimensional matrix $X = \left (\mathbf{x}_1 \dots \mathbf{x}_M \right )$, let $\permmat{X}{\pi}$ denote the sample-wise permutation under $\pi$. Thus,
\begin{align*}
    \permmat{X}{\pi} := \left (\mathbf{x}_{\pi(1)} \dots \mathbf{x}_{\pi(M)}\right ),
\end{align*}
with $K \in \mathbb{N}_{>0}$ the number of features. This means that the matrix $X$ is reordered by row.

\subsubsection*{Properties of permutations}
Next, we outline some properties of $S_{M}$ that are used in the proof of the main result. $S_{M}$ is a group with the composition of functions as group operator (denoted by $\circ$), thus the group axioms must hold \citep{Dixon,Artin}. This means that there exists an \emph{identity element} ${\text{id} \in S_M}$ such that for all ${\pi \in S_{M}}$:\begin{align*}
    \text{id}\circ \pi = \pi = \pi \circ \text{id} .
\end{align*}

Furthermore, for every ${\pi \in S_{M}}$, there exists a unique \emph{inverse element} ${\pi^{-1}\in S_{M}}$ such that
\begin{align*}
    \pi \circ \pi^{-1} = \text{id} = \pi^{-1} \circ \pi. %
\end{align*}
Thus, for each permutation, there exists an inverse permutation that reverses the change of order of the permutation, which is used in Section~\ref{sec: proof}. As each inverse is unique and also contained in $S_{M}$, it follows that
\begin{align}
    \{\pi \in S_{M}\} = \{\pi^{-1} : \pi\in S_{M}\}, \label{eq: set inverse permutation}
\end{align}
which means that the set of all permutations is the same as the set of all inverses of these permutations. Thus, taking an expectation over all permutations in $S_{M}$ is the same as taking the expectation over all inverse permutations of permutations in $S_{M}$. This is used in the proof of the main result in Section~\ref{sec: proof}.

\section{Essential conditions} \label{sec: essential conditions}
To prove that the optimal Dutch Draw classifier yields the `optimal' baseline, we first have to define `optimality'. When is a baseline considered to be optimal? To determine this, the following two questions must be answered: \begin{enumerate*}[label=(\arabic*)]
     \item which methods do we compare and
     \item how do we compare them?
    \end{enumerate*}
    To this end, we define the notion of \emph{input-independent} classifiers, \emph{positional-invariant} measures, and \emph{permutation-optimality}.

\subsection{Input-independent classifier}\label{sec: input-independent classifier}
Any binary classifier can be used as a baseline. However, any good standardized baseline should be \emph{general}, \emph{simple}, and \emph{informative} \Citep{Dutchdraw}. Thus, it needs to be applicable to any domain, quick to train and clearly still beatable. To this end, we investigate all models that do not take any feature values into account, as they meet these three requirements. Without considering feature values, they can be applied to any domain. Furthermore, they do not require any training, because they cannot learn the relationship between the feature values and the corresponding labels. This makes them also clearly still beatable, as any newly developed model should leverage the information from the feature values to make better predictions.

A binary classifier ${h_M\in \mathcal{H}_M}$ is called \emph{input-independent} if for all feature spaces ${\mathcal{X}_1^M}$, ${\mathcal{X}_2^M}$ and for all feature values ${\mathbf{X}_i \in \mathcal{X}_1^M}$ and ${\mathbf{X}_j\in \mathcal{X}_2^M}$ it holds that $h_M(\mathbf{X}_i)$ and $h_M(\mathbf{X}_j)$ are identically distributed. In other words,
\begin{align*}
    h_M(\mathbf{X}_i) \dequal h_M(\mathbf{X}_j) \dequal: h_M(\cdot), %
\end{align*}
where the notation of $h_M(\cdot)$ is chosen to visualize that the classifier $h_M$ is not dependent on the input. By this definition, an input-independent classifier is not dependent on feature values or even the feature domains. Let ${\hmii = \{h_M\in \mathcal{H}_M: h_M \text{ is input-independent}\}}$ be the set of all input-independent binary classifiers. A newly developed model, that was optimized using the same performance measure, should always beat the performance of an \emph{input-independent} model, as it gains information from the feature values. Otherwise, the model was not able to exploit this extra information to make better predictions.

\subsubsection*{Example of an input-independent classifier}
The \emph{coin flip} (see Section~\ref{sec: example binary classifier}) is by definition input-independent. The feature values have no influence on the probability distribution of the coin. Thus, for any $(x_1,\dots, x_M) \in \mathcal{X}^M$,
\begin{align*}
    \left ( \coinn(x_1), \dots, \coinn(x_M)  \right ) = \left ( \coinn(\cdot), \dots, \coinn(\cdot)  \right ) .
\end{align*}

\subsection{Positional-invariant measure}\label{sec: positional-invariant measure}
To assess the performance of a method, a measure needs to be chosen. Reasonably, the order of the samples should not change the outcome of this measure. If a measure has this property, we call it \emph{positional-invariant}. More formally, a measure $\mu$ is \emph{positional-invariant} if for every permutation ${\pi \in S_{M}}$ and for all ${h_M(\mathbf{X}),  \mathbf{Y} \in \{0,1\}^M}$ it holds that:
\begin{align}
    \mu\left (h_M(\mathbf{X}) , \mathbf{Y}\right ) = \mu(\permvec{h_M(\mathbf{X})}{\pi}, \permvec{ \mathbf{Y}}{\pi}). \label{eq: position-invariant}
\end{align}
This means that any reordering of the coupled \emph{predicted} and \emph{actual} labels does not affect the performance score.

This is not a hard restriction, as most measures have this property. Note for example that the number of \emph{true positives} (TP), \emph{true negatives} (TN), \emph{false positives} (FP), and \emph{false negatives} (FN) are all \emph{positional-invariant}. Most commonly used measures are a function of these four measures \citep{Sokolova}, making them also \emph{positional-invariant}.

\subsubsection*{Example of a non-positional-invariant measure}
Nonetheless, it is possible to define measures that are not \emph{positional-invariant}. For example, take the measure
\begin{align*}
    \lambda: \{0,1\} ^M \times \{0,1\} ^M \to \RR, \left (a=(a_1, \dots, a_M),b\right ) \mapsto a_1,
\end{align*}
which is dependent on the first position of the prediction, as \begin{align*}
    \lambda \left (\begin{pmatrix}
        0 \\ 1
    \end{pmatrix}, \begin{pmatrix}
        1 \\ 0
    \end{pmatrix} \right ) &= 0,\\
    \lambda \left (\begin{pmatrix}
      1 \\ 0
    \end{pmatrix}, \begin{pmatrix}
        0 \\ 1
    \end{pmatrix} \right ) &= 1.
\end{align*}

\subsection{Defining optimality} \label{sec: defining optimality}
To find the `optimal' baseline, it is first essential to specify what `optimality' entails.

\subsubsection{Optimal classifier}
 A binary classifier does not need to have a deterministic outcome. Thus, due to \emph{stochasticity}, we consider a classifier to be \emph{optimal} if it minimizes/maximizes the expected performance out of all considered binary classifiers (i.e., $\mathcal{\tilde{H}}_M \subseteq \mathcal{H}_M$). Whether optimization means minimization or maximization depends on the objective of the problem. So:
\begin{align}
    & h^{\text{min}}_M \in \argmin_{h_M \in \mathcal{\tilde{H}}_M} \left \{\EE_{h_M(\mathbf{X})} \left [\mu(h_M(\mathbf{X}) , \mathbf{Y}) \right ]   \right \}, \label{eq: optimal property min} \\
    & h^{\text{max}}_M \in \argmax_{h_M \in \mathcal{\tilde{H}}_M} \left \{\EE_{h_M(\mathbf{X})} \left [\mu(h_M(\mathbf{X}) , \mathbf{Y}) \right ]   \right \}. \label{eq: optimal property max}
\end{align}
For example, when the goal is to maximize the accuracy, then $h^{\text{max}}_M$ is an optimal baseline out of all other baselines in $ \mathcal{\tilde{H}}_M$. Note that there could be multiple different optimal baselines.

\subsubsection{Trivial optimal solution}
However, this definition of `optimality' leads to a trivial optimal solution, when we consider all \emph{input-independent} classifiers ($\mathcal{\tilde{H}}_M = \hmii$). Take the deterministic classifier $$\tilde{h}_M^{\text{max}}(\cdot)  := \hat{\mathbf{Y}}_{\text{max}}  \in \argmax_{\hat{\mathbf{Y}} \in \{0,1\}^M}\mu (\hat{\mathbf{Y}}, \mathbf{Y}),$$
which always predicts a vector $\hat{\mathbf{Y}}_{\text{max}}$ that maximizes the measure $\mu$. Note that $\tilde{h}_M^{\text{max}}$ is clearly \emph{input-independent} (see Section~\ref{sec: input-independent classifier}), thus $\tilde{h}_M^{\text{max}} \in \hmii$. Furthermore, it holds that
\begin{align*}
    \max_{h_M \in \mathcal{\tilde{H}}_M} \left \{\EE_{h_M(\mathbf{X})} \left [\mu(h_M(\mathbf{X}) , \mathbf{Y}) \right ]   \right \} & \leq \max_{\hat{\mathbf{Y}} \in \{0,1\}^M}\mu (\hat{\mathbf{Y}}, \mathbf{Y}) \\
    &=\EE_{\tilde{h}_M^{\text{max}}(\cdot)} \left [\mu(\tilde{h}_M^{\text{max}}(\cdot) , \mathbf{Y}) \right ]
    .
\end{align*}
In other words, the expected performance of $\tilde{h}_M^{\text{max}}$ is always higher or equal compared to any other classifier. Thus, $\tilde{h}_M^{\text{max}}$ is considered to be \emph{optimal} (see Equation~\eqref{eq: optimal property max}). The same holds for minimization with $$\tilde{h}_M^{\text{min}}(\cdot) := \hat{\mathbf{Y}}_{\text{min}} \in \argmin_{\hat{\mathbf{Y}} \in \{0,1\}^M}\mu (\hat{\mathbf{Y}}, \mathbf{Y}).$$ Essentially, a perfect prediction can always be made by an \emph{input-independent} classifier, using the \emph{actual} labels and the performance measure. Consider for example the commonly used performance measure: \emph{accuracy}, which is maximized if the prediction ${\hat{\mathbf{Y}} = \mathbf{Y}}$. A classifier $\tilde{h}_M^{\text{max}}$ that always predicts $\mathbf{Y}$, is thus \emph{optimal} for these given labels. This shows that an extension to the definition of `optimality' should be considered.

\subsubsection{Permutation-optimality}\label{sec: permutation-optimality}
The \emph{optimal} property (see Equations~\eqref{eq: optimal property min} and~\eqref{eq: optimal property max}) is not very insightful when we consider all deterministic classifiers, as the perfect prediction is always made by one of them. Similarly, a broken clock gives the correct time twice a day, but should not be used to determine the time. Therefore, we introduce a new optimality condition called \emph{permutation-optimality}.

It is often assumed that the test set is randomly shuffled. Therefore, we introduce the notion of \emph{permutation-optimality}. Instead of being optimal for the distinct order that the feature values and corresponding labels are given in, now all permutations of the samples are considered. A classifier is \emph{permutation-optimal} if it minimizes/maximizes the \emph{expected performance} for a random permutation of the test set out of all considered binary classifiers ($\mathcal{\tilde{H}}_M$). Thus,
\begin{align}
    & h^{\text{min}}_M \in \argmin_{h_M \in \mathcal{\tilde{H}}_M} \left \{\EE_{\pi \UD{S_{M}}}\left [\EE_{h_M(\permmat{\mathbf{X}}{\pi})} \left [\mu(h_M(\permmat{\mathbf{X}}{\pi}) , \permvec{\mathbf{Y}}{\pi}) \right ] \right ]  \right \}  \label{eq: definition shuffle optimal argmin},\\
    & h^{\text{max}}_M \in \argmax_{h_M \in \mathcal{\tilde{H}}_M} \left \{\EE_{\pi \UD{S_{M}}}\left [\EE_{h_M(\permmat{\mathbf{X}}{\pi})} \left [\mu(h_M(\permmat{\mathbf{X}}{\pi}) , \permvec{\mathbf{Y}}{\pi}) \right ] \right ]  \right \} \label{eq: definition shuffle optimal argmax}.
 \end{align}

\section{Dutch Draw classifier} \label{sec: dutch draw classifier}
A Dutch Draw classifier is defined in \Citep{Dutchdraw} for ${\theta \in [0,1]}$, as
\begin{align}
    &\TAC{\theta}(\mathbf{X}) := \left(\mathbf{1}_{E}(i)\right)_{i \in \{1, \dots M\}} \text{ with } E \subseteq \{1, \dots M\} \nonumber \\ & \qquad
    \text{ drawn \uar{}} \text{ such that } \vert E \vert = \MT. \label{eq: original definition dutchdraw}
\end{align}
In other words, the classifier draws \uar{} a subset $E$ of size $\MT$ out of all samples, which it then labels as 1, while the rest is labeled 0. In this section, we introduce an alternative definition, that is used in the main proof, and show that all Dutch Draw classifiers are \emph{input-independent}.

\subsection{Alternative definition} \label{sec: alternative definition dutch draw classifier}
Instead of the definition in Equation~\eqref{eq: original definition dutchdraw}, we introduce an alternative definition for the Dutch Draw classifiers to simplify the proof of the main result. Given a binary vector $(y_1, \dots, y_M) \in \{0,1\}^M$  of length $M$, note that the number of ones it contains can be counted by taking the sum $\sum_{i = 1}^M y_i$. Next, we define sets of binary vectors (of the same length) that contain the same number of ones. For all ${j\in \{0, \dots, M\}}$, define
\begin{align}
    \mathcal{Y}_j := \left \{\hat{\mathbf{Y}} = (y_1, \dots, y_M) \in \{0,1\}^M \text{ s.t. } \sum_{i = 1}^M y_i = j \right \}. \label{eq: definition yj groups}
\end{align}
In other words, $\mathcal{Y}_j$ contains all binary vectors of length $M$ with exactly $j$ ones and $M-j$ zeros. For example, for $M=4$ it holds that \begin{align*}
    \mathcal{Y}_0 &= \{(0,0,0,0)\},\\
    \mathcal{Y}_1 &= \{(0,0,0,1), (0,0,1,0), (0,1,0,0), (1,0,0,0)\},\\
    \mathcal{Y}_2 &= \{(0,0,1,1), (0,1,0,1), (0,1,1,0), (1,0,0,1), (1,0,1,0), (1,1,0,0) \},\\
    \mathcal{Y}_3 &= \{(0,1,1,1), (1,0,1,1), (1,1,0,1), (1,1,1,0) \},\\
    \mathcal{Y}_4 &= \{(1,1,1,1)\}.
\end{align*}
A Dutch Draw classifier selects \uar{} $E$ out of $M$ samples and labels these as one, and the rest zero. Note that this is the same as taking \uar{} a vector from $\mathcal{Y}_E$. To simplify notation, let $\mathcal{U}(A)$ denote the uniform distribution over a finite set $A$. Thus, when $X\sim\mathcal{U}(A)$ it must hold that $\PP(X = a)= \frac{1}{\vert A\vert }$ for each $a\in A.$ Now, a Dutch Draw classifier $\TAC{\theta}$ can be rewritten as
\begin{align}
    \TAC{\theta}(\mathbf{X}) &:= \hat{\mathbf{Y}} \text{ with } \hat{\mathbf{Y}} \UD{\mathcal{Y}_{\MT}}. \label{eq: alternative definition dutch draw}
\end{align}
Put differently, a Dutch Draw classifier $\TAC{\theta}$ chooses \uar{} a vector with exactly $\MT$ ones as prediction out of all vectors with $\MT$ ones ($\mathcal{Y}_{\MT}$). This alternative definition simplifies the proof of the main result.

\subsection{Input-independence} \label{sec: dutch draw classifier input-independent}
Next, we discuss why all Dutch Draw classifiers are \emph{input-independent} (see Section~\ref{sec: input-independent classifier}). Note that a Dutch Draw classifier $\TAC{\theta}$ is independent of feature values, as it is only dependent on $\theta$ and $M$, see Equation~\eqref{eq: alternative definition dutch draw}. In other words, any Dutch Draw classifier is by definition \emph{input-independent}. Instead of $\TAC{\theta}(\mathbf{X})$, we can therefore write $\TAC{\theta}(\cdot)$. To conclude, for every ${\theta \in [0,1]}$ it holds that ${\TAC{\theta}(\cdot) \in \hmii}$, which is the set of all input-independent binary classifiers.

\subsection{Optimal Dutch Draw classifier} \label{sec: optimal dutch draw classifier}
The optimal Dutch Draw classifier $\TAC{\theta_{\text{opt}}}$ is determined by minimizing/maximizing the expected performance for the parameter $\theta$ out of all allowed parameter values $\Theta$ \Citep{Dutchdraw}. Note that some measures are undefined for certain predictions, thus $\Theta$ is not always equal to $[0,1]$. Take e.g., the measure \emph{precision} \citep{Tharwat}, which is defined as the number of true positives divided by the total number of predicted positives. Therefore, if no positives are predicted, the measure becomes undefined (division by zero). By adapting each measure according to Section~\ref{sec: undefined measure}, all undefined cases are resolved and $\Theta = [0,1]$ always holds.

Using the alternative definition of the Dutch Draw classifier (see Equation~\eqref{eq: alternative definition dutch draw}), we obtain:
\begin{align}
    \theta^*_{\text{min}} &\in \argmin _{\theta \in [0,1]} \left \{ \EE_{\hat{\mathbf{Y}} \UD{\mathcal{Y}_{\MT}}} \left [\mu (\hat{\mathbf{Y}}, \mathbf{Y}) \right ]\right \}, \label{eq: chosen min dutch draw} \\
    \theta^*_{\text{max}} &\in \argmax _{\theta \in [0,1]} \left \{ \EE_{\hat{\mathbf{Y}} \UD{\mathcal{Y}_{\MT}}} \left [\mu (\hat{\mathbf{Y}}, \mathbf{Y}) \right ]\right \}. \label{eq: chosen max dutch draw}
\end{align}
Depending on the objective, either $\TAC{\theta^*_{\text{min}}}$ or $\TAC{\theta^*_{\text{max}}}$ is an optimal Dutch Draw classifier.

\section{Theorem and proof}\label{sec: proof}
After defining \emph{input-independence} (Section~\ref{sec: input-independent classifier}), \emph{positional-invariance} (Section~\ref{sec: positional-invariant measure}), \emph{permutation-optimality} (Section~\ref{sec: permutation-optimality}), and introducing an alternative formulation for the Dutch Draw classifier, all ingredients for the following theorem are present.
\begin{theorem}[Main result]
    The optimal Dutch Draw classifier $\TAC{\theta_{\text{opt}}}$ is \emph{permutation-optimal} out of all \emph{input-independent} classifiers ($\hmii$), for any \emph{positional-invariant} measure $\mu$. In other words: \begin{align}
        &\TAC{ \theta^*_{\text{min}}}  \in \argmin_{h_M \in \hmii} \left \{\EE_{\pi \UD{S_{M}}}\left [\EE_{h_M(\permmat{\mathbf{X}}{\pi})} \left [\mu(h_M(\permmat{\mathbf{X}}{\pi}) , \permvec{\mathbf{Y}}{\pi}) \right ] \right ]  \right \}, \label{eq: theorem shuffle optimal argmin} \\
        & \TAC{ \theta^*_{\text{max}}} \in \argmax_{h_M \in \hmii} \left \{\EE_{\pi \UD{S_{M}}}\left [\EE_{h_M(\permmat{\mathbf{X}}{\pi})} \left [\mu(h_M(\permmat{\mathbf{X}}{\pi}) , \permvec{\mathbf{Y}}{\pi}) \right ] \right ]  \right \}. \label{eq: theorem shuffle optimal argmax}
     \end{align}
\end{theorem}
This means that the optimal Dutch Draw classifier is the best general, simple, and informative baseline.

\begin{proof}
Let ${h_M\in \hmii}$ be an \emph{input-independent} classifier and let $\mu$ be a \emph{positional-invariant} measure, the classifier is \emph{permutation-optimal} if it minimizes/maximizes the expected performance under a random permutation of the test set out of all \emph{input-independent} classifiers (see Equations~\eqref{eq: definition shuffle optimal argmin} and~\eqref{eq: definition shuffle optimal argmax}).

For any \emph{input-independent} classifier $h_M$, it holds that

\begin{align}
    \EE_{h_M(\permmat{\mathbf{X}}{\pi})} \left [ \mu(h_M(\permmat{\mathbf{X}}{\pi}) , \permvec{\mathbf{Y}}{\pi}) \right ] &= \EE_{h_M(\cdot)} \left [ \mu(h_M(\cdot) , \permvec{\mathbf{Y}}{\pi}) \right ]. \label{eq: proof step 1}
\end{align}
The input $\permmat{\mathbf{X}}{\pi}$ is not relevant for the classification, and can thus be omitted.

In total, there are $2^M$ unique possible predictions in $\{0,1\}^M$. Denote these distinct vectors by $\hat{\mathbf{Y}}_1, \dots, \hat{\mathbf{Y}}_{2^M}$ such that $\bigcup_{i = 1}^{2^M} \hat{\mathbf{Y}}_{\mathbf{i}} =\{0,1\}^M.$ Next, the expectation in Equation~\eqref{eq: proof step 1} can be written out by:
\begin{align}
    \EE_{h_M(\cdot)} \left [ \mu(h_M(\cdot) , \permvec{\mathbf{Y}}{\pi}) \right ] &=  \sum_{i = 1}^{2^M} \PP(h_M(\cdot) = \hat{\mathbf{Y}}_{\mathbf{i}}) \cdot  \mu(\hat{\mathbf{Y}}_{\mathbf{i}}, \permvec{\mathbf{Y}}{\pi}). \label{eq: proof step 2}
\end{align}
As we need to proof \emph{permutation-optimality}, we have to take the expectation of Equation~\eqref{eq: proof step 2} over all permutations. Using linearity of expectation gives:
\begin{align}
  &\EE_{\pi \UD{S_{M}}}\left [ \sum_{i = 1}^{2^M} \PP(h_M(\cdot) = \hat{\mathbf{Y}}_{\mathbf{i}}) \cdot  \mu(\hat{\mathbf{Y}}_{\mathbf{i}}, \permvec{\mathbf{Y}}{\pi}) \right ] \nonumber \\ & \qquad = \sum_{i = 1}^{2^M} \PP(h_M(\cdot) = \hat{\mathbf{Y}}_{\mathbf{i}}) \cdot\EE_{\pi \UD{S_{M}}}\left [\mu(\hat{\mathbf{Y}}_{\mathbf{i}}, \permvec{\mathbf{Y}}{\pi}) \right ]. \label{eq: proof step 3}
\end{align}
Instead of taking the expectation of a sum, we now take the sum of expectations.

The measure $\mu$ is \emph{positional-invariant}, thus using Equation~\eqref{eq: position-invariant} gives \begin{align}
    \mu(\hat{\mathbf{Y}}_{\mathbf{i}}, \permvec{\mathbf{Y}}{\pi}) = \mu(\permvec{(\hat{\mathbf{Y}}_{\mathbf{i}})}{\pi^{-1}}, \permvec{(\permvec{\mathbf{Y}}{\pi})}{\pi^{-1}})= \mu(\permvec{(\hat{\mathbf{Y}}_{\mathbf{i}})}{\pi^{-1}}, \mathbf{Y}). \label{eq: proof step 3.5}
\end{align}
Applying a permutation does not change a \emph{positional-invariant} measure $\mu$. In this case, we apply the inverse permutation $\pi^{-1}$ to retrieve $\mathbf{Y}$.

Because of Equation~\eqref{eq: proof step 3.5}, it therefore also holds that
\begin{align}
    \EE_{\pi \UD{S_{M}}}\left [\mu(\hat{\mathbf{Y}}_{\mathbf{i}}, \permvec{\mathbf{Y}}{\pi}) \right ] &= \EE_{\pi \UD{S_{M}}}\left [\mu(\permvec{(\hat{\mathbf{Y}}_{\mathbf{i}})}{\pi^{-1}}, \mathbf{Y}) \right ]. \label{eq: proof step 4}
\end{align}
Equation~\eqref{eq: set inverse permutation} shows that the set of all \emph{inverse permutations} is the same as the set of all \emph{permutations}. Given that the permutations are drawn \uar{}, taking the expectation over all the \emph{inverse permutations} is the same as taking the expectation over all \emph{permutations}. When permutation $\pi$ is drawn \uar{}, it namely holds that ${\PP(\pi = s) = \PP(\pi = s^{-1}) = \frac{1}{\vert S_{M} \vert }}$ for all $s\in S_{M}.$ Therefore,
\begin{align}
    \EE_{\pi \UD{S_{M}}}\left [\mu(\permvec{(\hat{\mathbf{Y}}_{\mathbf{i}})}{\pi^{-1}}, \mathbf{Y}) \right ] &= \sum_{s \in S_{M}}\left (\mu(\permvec{(\hat{\mathbf{Y}}_{\mathbf{i}})}{s^{-1}}, \mathbf{Y}) \cdot \PP(\pi = s) \right ) \nonumber \\
    &= \sum_{s \in S_{M}}\left (\mu(\permvec{(\hat{\mathbf{Y}}_{\mathbf{i}})}{s^{-1}}, \mathbf{Y}) \cdot \PP(\pi = s^{-1}) \right ) \nonumber \\
    &= \EE_{\pi \UD{S_{M}}}\left [\mu(\permvec{(\hat{\mathbf{Y}}_{\mathbf{i}})}{\pi}, \mathbf{Y}) \right ]. \label{eq: proof step 5}
\end{align}
Thus, $\pi^{-1}$ can be replaced with $\pi$ in Equation~\eqref{eq: proof step 4}.

Recall that $\mathcal{Y}_j$ is the set of all binary vectors of length $M$ with $j$ ones (see Equation~\eqref{eq: definition yj groups}). Furthermore, note that applying a \uar{} chosen permutation $\pi\in S_{M}$ on ${\hat{\mathbf{Y}}_{\mathbf{i}} \in \mathcal{Y}_j}$ is the same as selecting \uar{} ${\hat{\mathbf{Y}}\in \mathcal{Y}_j}$ as outcome, because for every ${\hat{\mathbf{Y}}_{\mathbf{\star}} \in \mathcal{Y}_j}$ it holds that \begin{align*}
    \PP\left( \permvec{(\hat{\mathbf{Y}}_{\mathbf{i}})}{\pi} = \hat{\mathbf{Y}}_{\mathbf{\star}} \right)  = \frac{1}{\vert \mathcal{Y}_j \vert } \text{ with } \pi\UD{S_{M}},
\end{align*}
and
\begin{align*}
    \PP\left(  \hat{\mathbf{Y}} = \hat{\mathbf{Y}}_{\mathbf{\star}} \right)  = \frac{1}{\vert \mathcal{Y}_j \vert } \text{ with } \hat{\mathbf{Y}}\UD{\mathcal{Y}_j}.
\end{align*}
Therefore, we can rewrite the expectation ${\EE_{\pi \UD{S_{M}}}\left [\cdot \right ]}$ over all permutations into an expectation over a \uar{} drawn vector with the same number of ones, by
\begin{align}
   \EE_{\pi \UD{S_{M}}}\left [\mu(\permvec{(\hat{\mathbf{Y}}_{\mathbf{i}})}{\pi}, \mathbf{Y}) \right ] &= \EE_{\hat{\mathbf{Y}} \UD{\mathcal{Y}_j}: \hat{\mathbf{Y}}_{\mathbf{i}}\in \mathcal{Y}_j} \left [ \mu (\hat{\mathbf{Y}}, \mathbf{Y}) \right ]. \label{eq: proof step 6}
\end{align}
Using Equations~\eqref{eq: proof step 4}, \eqref{eq: proof step 5}, and \eqref{eq: proof step 6} in combination with Equation~\eqref{eq: proof step 3} gives
\begin{align*}
    &\sum_{i = 1}^{2^M} \PP(h_M(\cdot) = \hat{\mathbf{Y}}_{\mathbf{i}}) \cdot\EE_{\pi \UD{S_{M}}}\left [\mu(\hat{\mathbf{Y}}_{\mathbf{i}}, \permvec{\mathbf{Y}}{\pi}) \right ] \nonumber \\ & \qquad = \sum_{i = 1}^{2^M} \PP(h_M(\cdot) = \hat{\mathbf{Y}}_{\mathbf{i}}) \cdot \EE_{\hat{\mathbf{Y}} \UD{\mathcal{Y}_j}: \hat{\mathbf{Y}}_{\mathbf{i}}\in \mathcal{Y}_j} \left [\mu (\hat{\mathbf{Y}}, \mathbf{Y}) \right ]. %
\end{align*}
We have now eliminated all permutations from the equation. Note that the expectation in the right-hand side is the same for each $\hat{\mathbf{Y}}_{\mathbf{i}} \in \mathcal{Y}_j$. In other words, the expectation is the same for two vectors, when they have the same number of ones. Grouping the vectors with the same number of ones, gives
\begin{align*}
    &\sum_{i = 1}^{2^M} \PP(h_M(\cdot) = \hat{\mathbf{Y}}_{\mathbf{i}}) \cdot \EE_{\hat{\mathbf{Y}} \UD{\mathcal{Y}_j}: \hat{\mathbf{Y}}_{\mathbf{i}}\in \mathcal{Y}_j} \left [\mu (\hat{\mathbf{Y}}, \mathbf{Y}) \right ] \nonumber  \\ & \qquad = \sum_{j=0}^{M} \PP(h_M(\cdot) \in \mathcal{Y}_j) \cdot \EE_{\hat{\mathbf{Y}} \UD{\mathcal{Y}_j}}\left [\mu (\hat{\mathbf{Y}}, \mathbf{Y}) \right ]. %
   \end{align*}
Instead of summing over all possible binary vectors $\hat{\mathbf{Y}}_{\mathbf{i}} \in \{0,1\}^M$, all vectors with the same number of ones are grouped together, as they have the same expectation. All probability mass of the grouped vectors is also added up. Note, that it is thus only relevant for a classifier in which group $\mathcal{Y}_j$ the prediction $h_M(\cdot)$ belongs.

For any ${j\in \{0, \dots, M\}}$ it holds that ${\EE_{\hat{\mathbf{Y}} \UD{\mathcal{Y}_j}} \left [\mu (\hat{\mathbf{Y}}, \mathbf{Y}) \right ]}$ is bounded by minimizing/maximizing over all possible values of $j$. Thus,
\begin{align}
    \EE_{\hat{\mathbf{Y}} \UD{\mathcal{Y}_j}} \left [\mu (\hat{\mathbf{Y}}, \mathbf{Y}) \right ] & \geq \min_{j' \in \{0, \dots, M\}}\EE_{\hat{\mathbf{Y}} \UD{\mathcal{Y}_{j'}}} \left [\mu (\hat{\mathbf{Y}}, \mathbf{Y}) \right ], \label{eq: proof step 7.5 min} \\
    \EE_{\hat{\mathbf{Y}} \UD{\mathcal{Y}_j}} \left [\mu (\hat{\mathbf{Y}}, \mathbf{Y}) \right ] & \leq \max_{j' \in \{0, \dots, M\}}\EE_{\hat{\mathbf{Y}} \UD{\mathcal{Y}_{j'}}} \left [\mu (\hat{\mathbf{Y}}, \mathbf{Y}) \right ]. \label{eq: proof step 7.5 max}
\end{align}
Observe that ${\sum_{j=0}^{M} \PP(h_M(\cdot) \in \mathcal{Y}_j) = 1}$ and $ {\PP(h_M(\cdot) \in \mathcal{Y}_j) \geq 0}$ hold for each $j$, therefore it follows using Equations~\eqref{eq: proof step 7.5 min} and~\eqref{eq: proof step 7.5 max} that \begin{align*}
    \sum_{j=0}^{M} \PP(h_M(\cdot) \in \mathcal{Y}_j) \cdot \EE_{\hat{\mathbf{Y}} \UD{\mathcal{Y}_j}} \left [\mu (\hat{\mathbf{Y}}, \mathbf{Y}) \right ] & \geq \min_{j' \in \{0, \dots, M\}}\EE_{\hat{\mathbf{Y}} \UD{\mathcal{Y}_{j'}}} \left [\mu (\hat{\mathbf{Y}}, \mathbf{Y}) \right ], \\ %
    \sum_{j=0}^{M} \PP(h_M(\cdot) \in \mathcal{Y}_j) \cdot \EE_{\hat{\mathbf{Y}} \UD{\mathcal{Y}_j}} \left [\mu (\hat{\mathbf{Y}}, \mathbf{Y}) \right ] &  \leq \max_{j' \in \{0, \dots, M\}}\EE_{\hat{\mathbf{Y}} \UD{\mathcal{Y}_{j'}}} \left [\mu (\hat{\mathbf{Y}}, \mathbf{Y}) \right ]. %
\end{align*}
Consequently, we have found a lower and upper bound for Equations~\eqref{eq: theorem shuffle optimal argmin} and~\eqref{eq: theorem shuffle optimal argmax}, respectively. Namely,
\begin{align}
    &\min_{h_M \in \hmii} \left \{\EE_{\pi \UD{S_{M}}}\left [\EE_{h_M(\permmat{\mathbf{X}}{\pi})} \left [\mu(h_M(\permmat{\mathbf{X}}{\pi}) , \permvec{\mathbf{Y}}{\pi}) \right ] \right ]  \right \} \geq \min_{j \in \{0, \dots, M\}} \left \{\EE_{\hat{\mathbf{Y}} \UD{\mathcal{Y}_j}}\mu (\hat{\mathbf{Y}}, \mathbf{Y})\right \} , \label{eq: lower bound}  \\
    & \max_{h_M \in \hmii} \left \{\EE_{\pi \UD{S_{M}}}\left [\EE_{h_M(\permmat{\mathbf{X}}{\pi})} \left [\mu(h_M(\permmat{\mathbf{X}}{\pi}) , \permvec{\mathbf{Y}}{\pi}) \right ] \right ]  \right \} \leq \max_{j \in \{0, \dots, M\}} \left \{\EE_{\hat{\mathbf{Y}} \UD{\mathcal{Y}_j}}\mu (\hat{\mathbf{Y}}, \mathbf{Y})\right \}. \label{eq: upper bound}
\end{align}
Equality only holds for any classifier $h_M\in \hmii$, when all probability mass is given to {${\argmin_{j \in \{0, \dots, M\}} \left \{\EE_{\hat{\mathbf{Y}} \UD{\mathcal{Y}_j}}\mu (\hat{\mathbf{Y}}, \mathbf{Y})\right \}}$} and ${\argmax_{j \in \{0, \dots, M\}} \left \{\EE_{\hat{\mathbf{Y}} \UD{\mathcal{Y}_j}}\mu (\hat{\mathbf{Y}}, \mathbf{Y})\right \}}$, respectively. In other words, the minimum can only be attained if
\begin{align}
    \sum_{j_{\text{min}} \in \argmin_{j \in \{0, \dots, M\}} \left \{\EE_{\hat{\mathbf{Y}} \UD{\mathcal{Y}_j}}\mu (\hat{\mathbf{Y}}, \mathbf{Y})\right \}}  \PP(h_M(\cdot) \in \mathcal{Y}_{j_{\text{min}}}) = 1, \label{eq: requirement minimum}
\end{align}
and the maximum only if
\begin{align}
    \sum_{j_{\text{max}} \in \argmax_{j \in \{0, \dots, M\}} \left \{\EE_{\hat{\mathbf{Y}} \UD{\mathcal{Y}_j}}\mu (\hat{\mathbf{Y}}, \mathbf{Y})\right \}}  \PP(h_M(\cdot) \in \mathcal{Y}_{j_{\text{max}}}) = 1. \label{eq: requirement maximum}
\end{align}
A classifier $h_M \in \hmii$ can therefore only attain the minimum/maximum if all predictions belong to a group $\mathcal{Y}_j$ or possibly multiple groups that all minimize/maximize the expectation (depending on the objective).

Remember that the Dutch Draw selects the optimal classifier based on Equations~\eqref{eq: chosen min dutch draw} and \eqref{eq: chosen max dutch draw},which leads to
\begin{align*}
    \lfloor M \cdot\theta^*_{\text{min}} \rceil &\in \argmin _{j \in \{0, \dots, M\}} \left \{ \EE_{\hat{\mathbf{Y}} \UD{\mathcal{Y}_{j}}} \left [\mu (\hat{\mathbf{Y}}, \mathbf{Y}) \right ]\right \},\\
    \lfloor M \cdot \theta^*_{\text{max}} \rceil &\in \argmax _{j \in \{0, \dots, M\}} \left \{ \EE_{\hat{\mathbf{Y}} \UD{\mathcal{Y}_{j}}} \left [\mu (\hat{\mathbf{Y}}, \mathbf{Y}) \right ]\right \}.\\
\end{align*}
Combining this with the alternative definition of the Dutch Draw (Equation~\eqref{eq: alternative definition dutch draw}) directly gives that
\begin{align*}
    \sum_{j_{\text{min}} \in \argmin_{j \in \{0, \dots, M\}} \left \{\EE_{\hat{\mathbf{Y}} \UD{\mathcal{Y}_j}}\mu (\hat{\mathbf{Y}}, \mathbf{Y})\right \}}  \PP(\TAC{ \theta^*_{\text{min}}}(\cdot) \in \mathcal{Y}_{j_{\text{min}}}) = 1, \\
    \sum_{j_{\text{max}} \in \argmax_{j \in \{0, \dots, M\}} \left \{\EE_{\hat{\mathbf{Y}} \UD{\mathcal{Y}_j}}\mu (\hat{\mathbf{Y}}, \mathbf{Y})\right \}}  \PP(\TAC{ \theta^*_{\text{max}}}(\cdot) \in \mathcal{Y}_{j_{\text{max}}}) = 1.
\end{align*}
This shows in combination with Equations~\eqref{eq: requirement minimum} and~\eqref{eq: requirement maximum} that the optimal Dutch Draw classifier \emph{actually} attains the bound given in Equations~\eqref{eq: lower bound} and~\eqref{eq: upper bound}. It now follows that,
\begin{align*}
    &\TAC{ \theta^*_{\text{min}}}  \in \argmin_{h_M \in \hmii} \left \{\EE_{\pi \UD{S_{M}}}\left [\EE_{h_M(\permmat{\mathbf{X}}{\pi})} \left [\mu(h_M(\permmat{\mathbf{X}}{\pi}) , \permvec{\mathbf{Y}}{\pi}) \right ] \right ]  \right \}, \\ %
    & \TAC{ \theta^*_{\text{max}}} \in \argmax_{h_M \in \hmii} \left \{\EE_{\pi \UD{S_{M}}}\left [\EE_{h_M(\permmat{\mathbf{X}}{\pi})} \left [\mu(h_M(\permmat{\mathbf{X}}{\pi}) , \permvec{\mathbf{Y}}{\pi}) \right ] \right ]  \right \}. %
 \end{align*}
Thus, we can conclude that the optimal Dutch Draw classifier attains the minimum/maximum expected performance and is therefore \emph{permutation-optimal} for all \emph{input-independent} classifiers with a \emph{positional-invariant} measure.
\end{proof}

\section{Discussion and conclusion}\label{sec: discussion and conclusion}
A baseline is crucial to assess the performance of a prediction model. However, there are infinitely many ways to devise a baseline method. As a necessary check in the development process, \Citet{Dutchdraw} plead for a supplementary baseline that is \emph{general}, \emph{simple}, and \emph{informative}. In this paper, we have therefore examined all baselines that are independent of feature values, which makes them general and relatively simple. Additionally, these baselines are also informative, as it should be considered a major warning sign when a newly developed model is outperformed by a model that does not take any feature values into account. In this paper, we have shown that, out of all \emph{input-independent} binary classifiers, the Dutch Draw baseline is \emph{permutation-optimal} for any \emph{positional-invariant} measure. Our findings improve the evaluation process of any new binary classification method, as we have proven that the Dutch Draw baseline is ideal to gauge the performance score of a newly developed model.

Next, we discuss two points that could be considered an `unfair' advantage for the Dutch Draw baseline. First of all, we have considered in this paper classifiers that predict $M$ labels \emph{simultaneously}. This gives classifiers a potential advantage over classifying each sample \emph{sequentially}, as e.g., exactly $k$ out of $M$ samples can be labeled positive. This can only be done sequentially when a classifier is allowed to track previous predictions or to change based on the number of classifications it has made. Even with this advantage, we believe that all \emph{input-independent} models still remain clearly beatable by a newly developed model.

Secondly, the Dutch Draw baseline can be derived for most commonly used measures without any additional knowledge about the number of positive labels $P$. Nonetheless, it was shown in \Citep{Dutchdraw} that the Dutch Draw baseline can only be calculated for the measure \emph{accuracy} when it is known if $P\geq M/2$ holds. If the distribution of the training set is the same as the test set, the training set can be used to determine whether $P\geq M/2$ is likely to hold. Furthermore, a domain expert could estimate whether it is likely that a dataset contains more positives than negatives. Take for example a cybersecurity dataset, where there are often significantly less harmful instances and more normal instances \citep{Wheelus}. There are thus many ways to estimate if $P\geq M/2$ holds. Nevertheless, even if the Dutch Draw baseline uses this information (only for the \emph{accuracy}), we believe that any newly developed model should still beat the Dutch Draw baseline, as it does not use any feature values to improve prediction.

Finally, we address future research opportunities. In this paper, we have only considered \emph{binary} classification. A natural extension would be to also consider \emph{multiclass} classification \citep{Grandini}. Is a strategy similar to the Dutch Draw optimal in this case? Can a closed-form expression of the optimal baseline be derived? We believe that the three introduced properties (namely, input-independent, positional-invariant, and permutation-optimal) are still relevant for the multiclass case. This could help identify what kind of classifier is considered to be optimal. \Citet{Dutchdraw} stated that the Dutch Draw baseline could be used to scale existing measures. This paper provides more motivation to scale measures with the Dutch Draw baseline and not by using any other \emph{input-independent} classifier. Yet, it could still be investigated how each measure should be scaled in order to maximize the explainability behind a performance score.



\section*{Disclosure statement}
The authors have no relevant financial or non-financial interests to disclose.

\section*{Funding}
No funding was received for conducting this study.

\section*{Availability of data}
No datasets were used in this research.

\printendnotes

\bibliography{Dutchdraw,Dixon,Cauchy,Artin,Sokolova,Pirouz,Longjie,Buckley,Tharwat,Grandini,Wheelus}



\end{document}